%% file: paper_template.tex
\documentclass[conference]{IEEEtran}
\usepackage{times}
\usepackage{graphicx}

\usepackage{subcaption}  

\usepackage{multicol}
\usepackage{colortbl}
\usepackage{booktabs}
\usepackage{comment}
\usepackage{times}
\usepackage{epsfig}
\usepackage{lipsum}
\usepackage{amsmath}
\usepackage{pifont}
\usepackage{amssymb}
\usepackage{arydshln}
\usepackage{color}
\usepackage[font=small, labelfont=bf]{caption}
\usepackage{url}
\usepackage{here}
\usepackage{balance}
\usepackage{xspace}
\usepackage{multirow}
\usepackage{bbding}
\usepackage[ruled,vlined]{algorithm2e}
\usepackage{stfloats}
\usepackage{wrapfig}
\usepackage{placeins}
\usepackage{makecell}
\usepackage[table,dvipsnames]{xcolor}
\usepackage{xcolor}
\usepackage{fontawesome5}

\usepackage[bookmarks=true, colorlinks=true]{hyperref}

\makeatletter
    \let\NAT@parse\undefined
\makeatother
\usepackage[square, numbers, sort&compress]{natbib}

\newcommand{\ours}[0]{\rowcolor{orange!10}}
\renewcommand{\baselinestretch}{1.0} 

\DeclareCaptionLabelFormat{brackets}{[#2]}
\input{includes}

\usepackage[numbers]{natbib}
\usepackage{multicol}
\usepackage[bookmarks=true]{hyperref}
\usepackage{cleveref}

\begin{document}

\title{AsyncVLA: An Asynchronous VLA for \\ Fast and Robust Navigation on the Edge}



\author{
Noriaki Hirose$^{1,2}$,
Catherine Glossop$^{1}$,
Dhruv Shah$^{3}$,
Sergey Levine$^{1}$\\
$^{1}$University of California, Berkeley \quad
$^{2}$Toyota Motor North America \quad
$^{3}$Princeton University
}


%

\makeatletter
\let\@oldmaketitle\@maketitle%
\renewcommand{\@maketitle}{\@oldmaketitle%
    \centering
    \vspace{3mm}
    \includegraphics[width=0.99\linewidth]{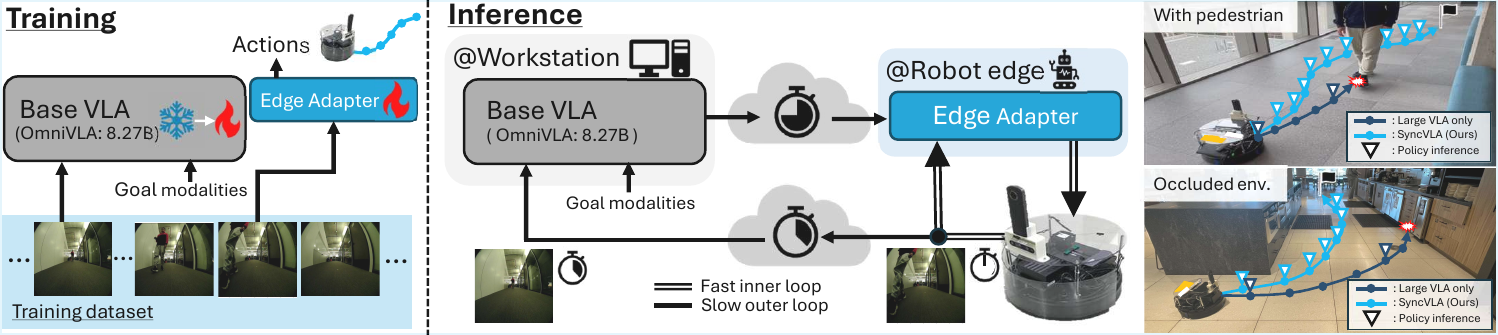}
    \vspace{0mm}
    \captionof{figure}{We train a fast and robust vision-based navigation policy by jointly optimizing a large robotic foundation model and a compact ViT in an end-to-end manner. During inference, we decouple the large base VLA and the \SmallModel{} to form a nested control system: the base VLA runs on a workstation to provide rich visual and language understanding, while the \SmallModel{} is deployed on the robot’s edge controller to rapidly adapt to the current environment and enable robust navigation.}
    \label{fig:pull}
}
\makeatother
\maketitle
\addtocounter{figure}{-1}
\thispagestyle{empty}
\pagestyle{empty}

\begin{abstract}
Robotic foundation models achieve strong generalization by leveraging internet-scale vision-language representations, but their massive computational cost creates a fundamental bottleneck: high inference latency. In dynamic environments, this latency breaks the control loop, rendering powerful models unsafe for real-time deployment. We propose AsyncVLA, an asynchronous control framework that decouples semantic reasoning from reactive execution. Inspired by hierarchical control, AsyncVLA runs a large foundation model on a remote workstation to provide high-level guidance, while a lightweight, onboard Edge Adapter continuously refines actions at high frequency. To bridge the domain gap between these asynchronous streams, we introduce an end-to-end finetuning protocol and a trajectory re-weighting strategy that prioritizes dynamic interactions. We evaluate our approach on real-world vision-based navigation tasks with communication delays up to 6 seconds. AsyncVLA achieves a 40\% higher success rate than state-of-the-art baselines, effectively bridging the gap between the semantic intelligence of large models and the reactivity required for edge robotics.
We present videos showcasing the performance and release our checkpoints and training code on our project page\footnote{\bf \href{https://asyncvla.github.io/}{\texttt{asyncvla.github.io}}}.
\end{abstract}

\IEEEpeerreviewmaketitle

\section{Introduction}
Recent progress in robot learning has been marked by the introduction of vision-language-action models (VLAs)~\citep{black2026pi0visionlanguageactionflowmodel, kim2024openvlaopensourcevisionlanguageactionmodel, hirose2025omnivla, belkhale2024minivla, shukor2025smolvlavisionlanguageactionmodelaffordable, wen2024tinyvla, li2025cogvlacognitionalignedvisionlanguageactionmodel}, models finetuned from large vision-language models (VLMs) on robot data. VLAs can inherit internet-scale knowledge from VLMs, resulting in vastly superior vision and language understanding compared to those trained from scratch. However, as we massively scale up the size of our robot policies, we sacrifice inference speed, which can impact performance~\cite{shukor2025smolvlavisionlanguageactionmodelaffordable,sendai2025leave,chen2025fast}. In particular, increased inference latency results in longer control cycles, which manifest as time delays in the control loop that can degrade overall system performance. This paper tackles the research question: {\it how can large robotic foundation models be deployed on the edge without being constrained by their computational cost?}

The challenges associated with large-scale foundation models are especially pronounced in mobile robots. Due to the limited field of view with ego-centric camera views, mobile robots cannot fully perceive their surroundings from a given observation~\cite{shah2023gnm,shahvint,hirose2023exaug}. With the presence of dynamic obstacles, such as pedestrians, this problem is further exacerbated~\cite{hirose2023sacson}.  Consequently, mobile robots must continuously update their observations while navigating and rapidly generate actions based on the most recent sensory inputs. As a result, the increasing computational burden associated with scaling robotic foundation models directly leads to performance degradation in real-world navigation tasks.
In addition, mobile robots often cannot be equipped with high-performance GPUs, making it necessary to execute foundation models on cloud-based or remote workstations~\cite{niwa2022spatio,11278012,erz_sdk}. In such settings, communication latency between the robot and the workstation introduces additional delays into the control loop, further degrading performance. Even when inference can be performed on board, the high power consumption of large foundation models significantly reduces battery life and severely limits operational duration.

Traditional control systems often use a hierarchical or cascaded architecture~\citep{Wang_2020, LEE2019358, vermillion2013stablehierarchicalmodelpredictive}. In navigation, for example, a global planner might serve as an outer loop providing high-level guidance, while a local planner functions as an inner loop, with the inner loop operating at a higher frequency to compensate for delays in the slower outer loop and incorporate local information. Similarly, in servo control for robotic joints or machine tools, multiple control loops, such as position, velocity, and force control, are organized hierarchically from outer to inner loops, with higher update rates in inner loops to suppress disturbances such as perturbations and friction~\cite{astrom1995pid}.

Analogously, we construct a control system for VLAs that consists of an outer loop driven by a large-scale robotic foundation model and an inner loop governed by a lightweight model, both trained for end-to-end control. We make several key design decisions to allow our framework to be robust to large delays while maintaining the expressivity of the outer-loop VLA. We adapt a \SmallModel{} inner-loop policy, 100x smaller than the VLA, that maintains high reactivity while effectively incorporating high-level visual and language understanding guidance from the outer-loop policy. The lightweight inner-loop policy can incorporate the most recent observation information, enabling rapid correction when obstacles are encountered. We train our method by up-weighting a dynamic subset of our training data and use a two-staged training process: we first train the \SmallModel{} and then train the outer- and inner-loop framework end-to-end, better aligning the two policies. 


Our primary contribution is \MethodName{}, a framework that enables the deployment of large foundation models on edge robots by structurally decoupling reasoning from actuation. To this end, we developed an \SmallModel{}, a lightweight (76M parameter) policy that runs \emph{on board} the robot, fusing real-time sensors with delayed, high-level guidance from a remote VLA (over 8B parameters). Crucially, we propose a training strategy that up-weights reactive trajectories—moments where the robot must deviate from the base policy to avoid dynamic obstacles—and an end-to-end finetuning protocol to align the asynchronous streams. Our real-world experiments demonstrate that AsyncVLA maintains robust performance even under control delays that fluctuate up to 6 seconds, outperforming synchronous baselines by over 40\% in dynamic navigation scenarios.
\section{Related work}
%
Our work builds on recently developed large robotic foundation models, specifically vision-language-action models (VLAs). We will first introduce recent work on VLAs, then prior efforts to improve inference speed and efficiently deploy these models, and finally explore recent work on dual or fast-slow architectures for VLA deployment.

\vspace{2mm}
\noindent \textbf{Vision-language-action models.}
With the advent of large and powerful vision-language models, vision-language-action models (VLAs), which finetune VLMs for robot control~\citep{black2026pi0visionlanguageactionflowmodel, kim2024openvlaopensourcevisionlanguageactionmodel, hirose2025omnivla, belkhale2024minivla, shukor2025smolvlavisionlanguageactionmodelaffordable, wen2024tinyvla, li2025cogvlacognitionalignedvisionlanguageactionmodel}, have become a widely studied approach for instilling robot policies with generalist knowledge and reasoning capabilities. Although these models range in size, with some under 1 billion parameters~\citep{shukor2025smolvlavisionlanguageactionmodelaffordable, belkhale2024minivla, zheng2025xvlasoftpromptedtransformerscalable, lin2025evo1lightweightvisionlanguageactionmodel}, large models, in the range of 4 to 100s of billions of parameters~\citep{black2026pi0visionlanguageactionflowmodel, kim2024openvlaopensourcevisionlanguageactionmodel, geminiroboticsteam2025geminiroboticsbringingai, hirose2025omnivla}, tend to be more performant~\citep{brohan2023rt2visionlanguageactionmodelstransfer, generalist2025gen0}, just as their VLM base models tend to be more performant~\citep{kaplan2020scalinglawsneurallanguage}. These larger models can have long inference times, impeding reactive real-time control. \MethodName{} decouples inference into the large, powerful VLA, run on a remote workstation, and an \SmallModel{} action head, which can run directly on the robot's onboard compute. The \SmallModel{} action head can leverage the local information from the most up-to-date observations of the environment and rapidly refine the semantically informed actions from the VLA. Not only does make the inference speed of the VLA effectively the inference speed of the \SmallModel{}, we can also compensate for large fluctuations in network latency, which a VLA run off-board would always have to contend with, and respond to dynamic changes in the environment. 

\vspace{2mm}
\noindent 
\textbf{Efficient inference for VLAs.}
Previous work has aimed to improve the efficiency of VLAs using several methods. Reducing model size has been a key method as it directly addresses the cause of slow inference, whether that be through training smaller models~\citep{shukor2025smolvlavisionlanguageactionmodelaffordable, belkhale2024minivla, budzianowski2025edgevlaefficientvisionlanguageactionmodels, williams2025litevlaefficientvisionlanguageaction}, or quantizing parameters to use fewer bits~\citep{williams2025litevlaefficientvisionlanguageaction, wang2025bitvla1bitvisionlanguageactionmodels, belkhale2024minivla}. However, reducing model size entails that we have a limited set of VLAs that may be practical for edge deployment, while quantization may have a limited impact on model size before performance starts to degrade. Alternatively, other approaches improve inference speed through fast action decoding~\citep{pope2022efficientlyscalingtransformerinference, frans2025stepdiffusionshortcutmodels}, or compensate for inference delays with asynchronous inference methods~\citep{black2025trainingtimeactionconditioningefficient, tang2025vlashrealtimevlasfuturestateaware, zhao2025vlarailrealtimeasynchronousinference}. Along this same line, many methods append a fast ``action expert'' or action head~\citep{black2026pi0visionlanguageactionflowmodel, wen2024tinyvla, wen2025dexvlavisionlanguagemodelplugin, wen2024diffusionvla} onto the model, and others additionally use pipelines to decide when it is necessary to run inference over the full backbone~\citep{lin2025onetwovlaunifiedvisionlanguageactionmodel, bhattacharjya2025averyadaptivevlmsplit}. While these approaches allow for efficient inference, in that the entire backbone does not need to be used each time inference is performed, allowing for fast inference loops with just the action expert, these methods do not decouple the architecture, allowing the on-board inference on the action head alone. \MethodName{} explicitly separates inference into a small on-board policy, which gets the most recent observation information and can be rapidly inferred, and an off-board VLA, which provides high-level visual and semantic guidance. Additionally, while these methods only consider their own inference delay, on the order of 100s of milliseconds, \MethodName{} is explicitly trained to compensate for networking delays, even up to several seconds, and preserves the performance of the base VLA policy. 

\vspace{2mm}
\noindent 
\textbf{Hierarchical VLAs.}
A recent class of approaches that enables faster inference of large VLAs by using dual or ``fast-slow'' architectures, inspired by the dual system theory~\citep{kahneman2011thinking}. These architectures separate the VLA into a large backbone that runs at a low frequency, and a small, reactive model that runs at a higher frequency. These methods often use a latent from the large VLA to bridge the representations learned by the large model to the small one~\citep{han2024dualprocessvlaefficient, zhang2025hirtenhancingroboticcontrol, cui2025openhelixshortsurveyempirical, bu2025synergisticgeneralizedefficientdualsystem, shentu2025llmsactionslatentcodes, chen2025fast}, while others formulate the small model as a correction head and instead modify the actions coming from the base VLA~\citep{sendai2025leave}.
\MethodName{} also employs a dual architecture, but instead projects the token space of the base VLA policy into a smaller token space for the \SmallModel{}. More critically, these works focus on manipulation with a static workstation, and none need to consider the added delay of network latency that comes from deploying their system over a WiFi network. As a result, these works aim to compensate for small inference delays, up to about 300 ms. Few works include dynamic tasks in their evaluation and, when they are included, they consider very slow motion (objects moving at 1 cm/s)~\citep{zhang2025hirtenhancingroboticcontrol} and predictable patterns such as straight lines or circles~\citep{cui2025openhelixshortsurveyempirical}. Therefore, the small ``fast'' models in these systems are mostly tasked with interpolating the actions from the ``slow'' model based on recent sensor information. \MethodName{} is designed to handle unpredictable, dynamic situations, such as people walking in front of a moving robot, while handling variable network latencies of around 5 seconds, in addition to inference delays. As a result, the \SmallModel{} must be able to effectively learn skills like dynamic obstacle avoidance while using the high-level guidance provided by the base VLA.  To this end, we expose the \SmallModel{} to large variations between the most recent observation and the observation processed by the VLA. First, we randomly select the VLA's observation from a long context while the \SmallModel{} always processes the most up-to-date observation, forcing the \SmallModel{} to learn to predict actions that correspond to the current state of the environment, rather than the stale information processed in the last VLA inference pass. Second, we up-weight the segments of our dataset where the robot interacts with dynamic obstacles, such as encountering pedestrians, so that it must learn to predict their behavior and respond accordingly. 


To our knowledge, \MethodName is the first method to deploy a dual architecture with a large VLA and a small \emph{edge} model for navigation while explicitly accounting for both inference and network delays. We evaluate our method in both static and dynamic environments on pose- and language-conditioned navigation tasks and specifically demonstrate the robustness of \MethodName{} to enforced latencies of up to 6 seconds.   
\section{An Asynchronous VLA for Navigation}
\label{sec:method}
%
We propose an asynchronous navigation system, \MethodName{}, that leverages the rich language and visual understanding of existing robotic foundation models, while providing for quick reactive control based on the most recent observations. To achieve such a reactive policy without sacrificing rich semantic understanding, \MethodName{} integrates three key design choices into a single system: (1) a lightweight action head, \SmallModel{}, (2) automatic data re-balancing to encourage reactive behavior, and (3) end-to-end training for better aligning the base VLA and the \SmallModel{}.
%
%
\begin{figure}[t]
  \vspace{0mm}
  \begin{center}
      \includegraphics[width=0.99\hsize]{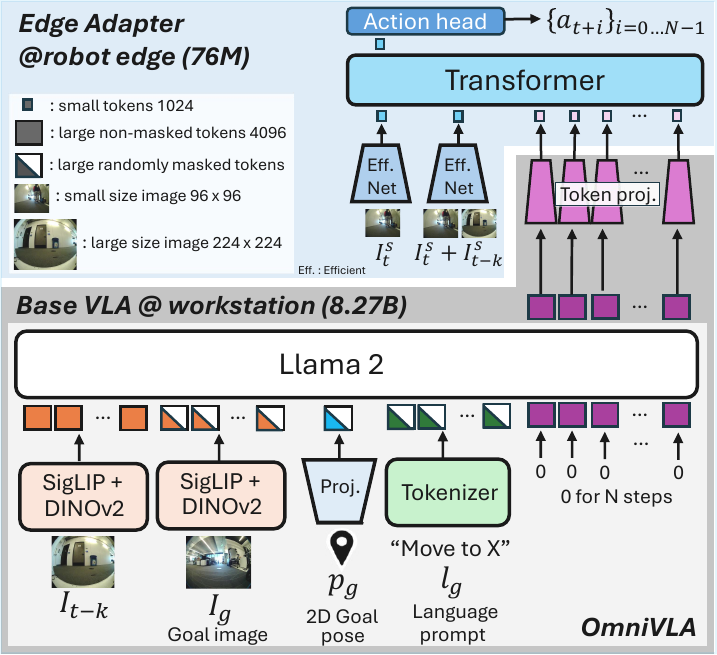}
  \end{center}
  \vspace{-2mm}
  \caption{\small {\bf Network architecture.} We train a \SmallModel{} on top of the large robotic foundation model, OmniVLA, for vision-based navigation. During inference, the \SmallModel{} runs on the robot's onboard controller at maximum speed to adjust the robot's behavior to the current environment, and the base VLA runs on the workstation to provide rich visual and language understanding.}

  \label{f:network}
  \vspace*{-4mm}
\end{figure}
\subsection{\SmallModel{} Architecture}
%
Robotic foundation models generate actions using rich language and visual understanding, but their large model size incurs substantial computational overhead, often requiring deployment on a remote workstation.
This setup introduces significant communication latency, reduces the achievable control frequency, and ultimately degrades task performance. With a latency of $k$ steps, by the time an action is executed by the robot, $k$ steps will have passed, so naively using a slow VLA will result in the action $a_t$ corresponding to the observation at time $t-k$, $I_{t-k}$, instead of the observation at time $t$, $I_t$.
To address this issue, we propose \MethodName, a policy architecture in Fig.~\ref{f:network} that builds upon an existing robotic foundation model, the base VLA, by augmenting it with a lightweight onboard policy(\SmallModel). In our implementation, we use OmniVLA~\cite{hirose2025omnivla} as the base VLA, which is further described in~\cref{sec:imp}.

Our \SmallModel{} adapts the actions of the base VLA to the information in the current observation, leveraging the semantic and visual information encoded in the base VLA's action token embeddings. To enable fast inference while leveraging the large model’s understanding as much as possible, we provide action token embeddings derived from the final-layer action features of the base VLA,
keeping \SmallModel{} lightweight with only 76M parameters. This allows \SmallModel{} to operate at a high frequency. However, the action token embeddings from the base VLA are temporally delayed, since the base model processes observations from $I_{t-k}$.
Note that the delay $k$ can fluctuate depending on network latency and workstation performance.

We address the challenge of preserving the semantic and visual information provided by the base VLA through its action token embeddings while accounting for the latency $k$ within the \SmallModel{}. This includes responding to dynamic changes in the environment that are not perceived at time $t-k$ and therefore are not included in the information provided in the action token embeddings. 
We design \SmallModel{} as a small ViT network (shown in the light blue area of Fig.~\ref{f:network}. In addition to the $8 \times 1024$ action token embeddings, which are projected from the base VLA's token space, we introduce two additional token embeddings of dimension $1024$. The first additional token embedding encodes the feature representation of the current image $I^s_t$, enabling the model to capture the robot’s immediate surroundings (see the leftmost input to the ``Transformer'' in Fig.~\ref{f:network}). The second token embedding is extracted from a concatenated six-channel image composed of $I^s_t$ and $I^s_{t-k}$, enabling the model to account for changes in robot pose and environmental dynamics occurring between $t-k$ and $t$ (see the second input from the left to the ``Transformer'' in Fig.~\ref{f:network}). In the \SmallModel{}, we resize the images to $96 \times 96$, denoted as $I^s$, which is less than one quarter of the resolution used by the base VLA. As a result, we can train the \SmallModel{} to be a fraction of the base VLA's size while preserving the rich information from the base VLA. 

To generate an action chunk for $N$ steps, $\{ a_{t+i}\}_{i=0 \ldots N-1}$, we attach a small action head on top of the transformer in \SmallModel{}. Here, $a_{t+i}$ represents the 2D pose at $t+i$ in the robot's coordinate frame at time $t$, similar to OmniVLA~\cite{hirose2025omnivla}, ViNT~\cite{shahvint} and MBRA~\cite{11278012}. Importantly, only the output token corresponding to $I_t$ is fed into the action head. This design conditions action generation on the current observation and prevents the \SmallModel{} from producing delayed actions influenced by the embeddings from the base VLA with $I_{t-k}$. Further details are described in Section \ref{sec:imp}.
\subsection{Up-weighting Reactive Trajectories} 
%
To make \MethodName{} a reactive policy, we require data sequences that capture reactive behaviors, such as collision avoidance or yielding to pedestrians. Such behaviors typically involve rapid changes in velocity or direction of the robot taking place within a single action chunk of the base VLA. However, such behaviors are relatively rare in the pre-training dataset, and training \SmallModel{} na\"{i}vely on this distribution is insufficient to learn the reactive behaviors we desire. To address this imbalance, we apply a data re-balancing approach that automatically identifies sequences containing sudden intra-chunk action changes and up-weights them during training.
%
%

Concretely, we extract from the training data two action chunk references, $A^t = \{ a^{\text{ref}, t}_{t+i} \}_{i=0 \ldots N-1}$ expressed in the robot coordinate frame at time $t$, and $A^{t-k} = \{ a^{\text{ref}, t-k}_{t-k+i} \}_{i=0 \ldots N-1}$ expressed in the robot coordinate frame at time $t-k$. Here, $A^t$ serves as the action reference for training \MethodName{}, while $A^{t-k}$ corresponds to the action reference for the observation $I_{t-k}$ of the base VLA, though it is \emph{not} used during training. If the robot maintains the same behavior (e.g., constant velocity) between $t-k$ and $t+N-1$, then $A^t = A^{t-k}$. Conversely, $A^t \neq A^{t-k}$ when the robot changes its behavior. We then compare these two action chunks to identify the data sequences that include changes in the robot's actions and up-weight the samples exhibiting large discrepancies. Specifically, a sample is up-weighted if the distance between the final pose of $A^t$ and that of $A^{t-k}$ exceeds a predefined threshold, $d_{\text{th}}$ such that $dist(a^{\text{ref}, t}_{t+N-1}, a^{\text{ref}, t-k}_{t-k+N-1}) > d_{\text{th}}$. 
\subsection{End-to-end Training}
We train \MethodName{} to learn a reactive \SmallModel{} that can dynamically respond to the most recent observation while leveraging the rich understanding of the large base VLA. Our base model, $\pi_b$ can be formulated as follows:
\begin{equation}
    \{a^{\text{emb}}_{t+i}\}_{i=0 \ldots N-1} = \pi_b^{\{\phi,\psi\}}(I_{t-k}, g),     
     \label{eq:large_model}
\end{equation}
where $\phi$ denotes the parameters of the token projector (see Sec.~\ref{sec:net_detail}), and $\psi$ denotes the parameters of the base VLA excluding $\phi$. $a^{\text{emb}}_{t+i}$ denotes the $i$-th action token embedding. $g$ denotes the goal, which could be specified in several modalities, such as an image, a 2D pose, or a language prompt. In Fig.~\ref{f:network}, these are represented as $I_g$, $p_g$, and $l_g$, respectively. Note that in our implementation, OmniVLA can be independently conditioned on either $I_g$, $p_g$, $l_g$, or a combination of these. In our experiments, we focus on the independent cases for $p_g$ and $l_g$.
We feed $\{a^{\text{emb}}_{t+i}\}_{i=0 \ldots N-1}$ into the following \SmallModel{}, $\pi_s$ and generate the action chunk $\{a_{t+i}\}_{i=0 \ldots N-1}$.
\begin{equation}
    \{a_{t+i}\}_{i=0 \ldots N-1} 
    = \pi_s^{\theta} (\{a^{\text{emb}}_{t+i}\}_{i=0 \ldots N-1},  I^s_t, I^s_{t-k}),     
     \label{eq:small_model}
\end{equation}
where $\theta$ denotes the parameters of the \SmallModel{}, 

We train our policy in two steps. At the beginning of the first step, we initialize $\psi$ with the base VLA's pretrained weights. Then, we train $\theta$ and $\phi$ from scratch, with $\phi$ frozen. This procedure allows $\theta$ and $\phi$ to acquire reasonable initial weights without degrading the base VLA's original performance.
%
%
%
In the second step, we fine-tune the entire policy end-to-end to better align the base VLA and the \SmallModel{}. 
In both training steps, we optimize the same objective using backpropagation.
\begin{align}
\min_{P} \; J(P) := J_{\text{im}}(P) + J_{\text{sm}}(P),
\label{eq:objective}
\end{align}
where $J_{\text{im}}$ denotes the imitation loss for the action chunk and $J_{\text{sm}}$ denotes a smoothing objective on the actions. 
In the first step, the parameter set is defined as $P = \{\theta, \phi\}$, while in the second step it is extended to $P = \{\theta, \phi, \psi\}$. Details of $J_{\text{im}}$ and $J_{\text{sm}}$ are provided in the appendix.

\section{Implementation and System Design}
\label{sec:imp}
We describe the implementation details for training \MethodName{}, as well as the system design for asynchronous inference on the edge.
%
\subsection{Network architecture}
\label{sec:net_detail}
We build \MethodName{} on top of OmniVLA, a vision-based robotic foundation model for navigation, which serves as our base VLA. OmniVLA is a navigation policy capable of generating actions conditioned on multiple goal modalities, including 2D goal poses, language instructions, and egocentric goal images. The base VLA interprets these diverse modalities at a high level using large-scale visual encoders (SigLIP~\cite{zhai2023sigmoid} and DINOv2~\cite{oquab2023dinov2}) and a language model (LLaMA2 7B~\cite{touvron2023llama}).

Since the downstream \SmallModel{} is responsible for action generation, we feed it the action token embeddings
of OmniVLA, which encode features that represent the actions but also encode the semantic and visual information interpreted by OmniVLA. However, the embeddings produced by OmniVLA have a dimensionality of 8x4x4096, corresponding to an action chunk size of 8, an action dimension of 4, and an embedding size of 4096. Directly feeding these high-dimensional embeddings into the \SmallModel{} would significantly increase the network size and computational cost in the robot's onboard controller.

To address this issue, we apply a token projector with two MLP ResNet blocks to each action token, compressing its dimensionality from 4×4096 to 1024 before feeding it into the \SmallModel{}. This compression enables efficient transmission of the generated token embeddings from the workstation to the robot’s onboard controller during inference. For further details on OmniVLA, please see the original paper~\cite{hirose2025omnivla}.

In addition to the aspects described in Sec.~\ref{sec:method} for the \SmallModel{}, we ensure efficient, low-latency computation by extracting image features using a lightweight visual encoder, EfficientNet-B0~\cite{pmlr-v97-tan19a}, and by resizing input images to $96 \times 96$, which is less than one quarter of the resolution used by the base VLA. The transformer output corresponding to the current image embedding is then passed to an action head composed of a four-layer MLP, which predicts an action chunk $\{a_{t+i}\}_{i=0 \ldots N-1}$ for $N$ steps.
\subsection{Training dataset}
We train \MethodName{} on a mixture of public datasets to promote strong generalization. Our training mixture consists of: (1) the GNM dataset, which aggregates six publicly available robotic datasets—RECON~\cite{shah2022rapid}, CoryHall~\cite{kahn2018self}, TartanDrive~\cite{triest2022tartandrive}, Seattle~\cite{shaban2022semantic}, SCAND~\cite{karnan2022socially}, and GO Stanford 4~\cite{hirose2019deep}; (2) the LeLaN dataset~\citep{hiroselelan}, which includes in-the-wild videos and language-conditioned navigation data; and (3) the SACSoN (HuRoN) dataset~\cite{hirose2023sacson}, which captures dynamic scenes involving pedestrians.
\subsection{Training details}
We set the action chunk size $N$ to 8 at a control frequency of 3~Hz, resulting in a temporal horizon of 2.4~seconds for all models. For each training batch, data are uniformly sampled from the GNM~\cite{shah2023gnm}, SACSoN~\cite{hirose2023sacson}, and LeLaN~\cite{hiroselelan} datasets to balance the dataset domains. For half of each batch, we apply our up-weighting method with a threshold of $d_{\text{th}} = 1.0$~m. Additionally, for SACSoN, we prioritize samples that include pedestrians using the SACSoN annotations to train robotic behaviors that safely navigate around them. Due to memory constraints that limit batch sizes even on multi-GPU servers, we employ gradient accumulation over multiple steps to stabilize training.

When training \MethodName{} initialized from OmniVLA checkpoints on five H200 GPUs, we use a per-GPU batch size of 18 and accumulate gradients over two steps, yielding an effective batch size of 180 ($=5\times18\times2$). To further improve training efficiency, we apply LoRA, which limits the number of trainable parameters to approximately 5\% of the model, enabling larger effective batch sizes while maintaining a balance between training speed and stability. LoRA is applied only to the OmniVLA-based model due to its large parameter count. All other hyperparameters, including the learning rate, language tokenization, normalization, and related settings, follow the defaults in the OmniVLA codebase~\cite{omnivla_code}.
\subsection{System integration for inference}
During inference, we decouple \MethodName{} into the base VLA and the \SmallModel{}. The base VLA is deployed on a workstation equipped with a high-performance GPU, while the \SmallModel{} is implemented on the robot’s onboard controller to process observations without additional latency. An overview of the system implementation is shown in Fig.~\ref{f:vizbot_system}.

Due to its lightweight architecture, the \SmallModel{} can be executed efficiently on the onboard controller, enabling rapid action generation that reacts to changes in the surrounding environment while leveraging the rich visual and semantic features
provided by the base VLA. In contrast, since the base VLA is deployed on a workstation and is not physically connected to the robot, the robot transmits its observed images to the workstation via WiFi for the base VLA inference pass. The action token embeddings produced by the token projector in the base VLA and its time stamp $t-k$ are then sent back to the robot controller over WiFi.

On the robot controller, when receiving the updated action token embeddings, both the embeddings and the delayed observation $I^s_{t-k}$ to the \SmallModel{} are updated. Since $I^s_{t-k}$ corresponds to a past image, the robot controller maintains a buffer of past images along with their timestamps, and selects the appropriate $I^s_{t-k}$ by matching the timestamp of the received action token embeddings with the buffered observations. Similar to ViNT~\cite{shahvint} and NoMaD~\cite{sridhar2024nomad}, since our action space is the 2D pose, we use a PD controller to generate the velocity commands, which enables the robot to follow the generated 2D poses as actions.
We illustrate this process in Algorithm~\ref{alg:asyncvla_async}.
\begin{algorithm}[t]
\small
\caption{AsyncVLA Inference with Asynchronous Control Frequencies}
\label{alg:asyncvla_async}

\KwIn{Base VLA $\pi_b$ (frequency $f_b$), \SmallModel{} $\pi_s$ (frequency $f_s$), token projector $\phi$}
\KwOut{Executed robot actions}

Deploy $\pi_b$ on a GPU-equipped workstation\;
Deploy $\pi_s$ on the robot onboard controller\;
Initialize image buffer $\mathcal{B} \leftarrow \emptyset$\;

\While{robot is operating}{
    \textbf{Onboard controller loop} (runs at $f_s$ ($> f_b$) Hz):\;
    Capture current observation $I_t$ with timestamp $t$\;
    Store $(I_t, t)$ in buffer $\mathcal{B}$\;
    Send $I_t$ and t to the workstation\;
    
    \If{new action token embeddings are received}{
        Retrieve delayed observation $I_{t-k}$ from $\mathcal{B}$ via timestamp matching\;
        Update \SmallModel{} action token embedding input\;
    }
    
    Compute action chunk $\{\Delta a_{t+i}\}_{i=0}^{N-1}$ using $\pi_s$\;
    Compute PD controller to generate vel. commands.\;
    Execute vel. commands.;
    
    \textbf{Workstation loop} (runs at $f_b$ Hz, asynchronously):\;
    Receive $I_{t-k}$ and $t-k$ from the robot controller\;
    Compute base-model inference using $\pi_b$ with $I_{t-k}$ and the given goal inputs ($l_g$, $p_g$, $l_g$)\;
    Send action token embeddings and timestamp $t-k$ to the robot controller\;
}
\end{algorithm}
\section{Experimental Setup}
\label{sec:evaluation_setup}
We first describe our experimental setup for evaluating \MethodName{} on real-world robot platforms. We then conduct extensive real-world evaluations and compare our method against state-of-the-art baselines.

\subsection{Navigation tasks}
We consider the following two navigation tasks:

\textit{{\bf Goal pose-conditioned navigation:}} 
When conditioned on 2D goal poses, our policy navigates the robot from its initial position to a specified target pose $p_g$. In our experiments, the distance between the initial position and the target pose ranges from 12 to 30 meters. We conduct experiments in diverse environments, including both indoor and outdoor settings. 

For evaluation, we select ten challenging, cluttered environments and perform two trials at different times in each environment. In at least one trial per environment, pedestrians are present and interact with the robot. These interactions are kept consistent across all evaluated methods to ensure fair and reproducible comparisons.
%
%
%

\textit{{\bf Language-conditioned navigation:}} 
We assess \MethodName{} using language commands that not only indicate the target destination but also describe the desired navigation behavior during execution.
Experiments are carried out in twelve distinct environments, including office spaces, kitchen areas, and entrance halls, with a diverse set of language instructions. Target locations are positioned between 5 and 20 meters from the robot’s starting pose.

To examine the advantages of leveraging OmniVLA, we evaluate performance under out-of-distribution (OOD) language commands that are not explicitly represented in the training data. Whereas the training set mainly consists of object-directed navigation instructions (e.g., “move toward X”), we design out-of-distribution (OOD) prompts that specify how the robot should navigate toward the goal in half of the evaluation trials (e.g., “move toward X along Y”). These prompts are constructed following the protocol of CAST~\cite{glossop2025cast}, and the robot’s behavior is evaluated based on its compliance with the provided instructions.
\begin{figure}[t]
  \begin{center}
      \includegraphics[width=0.99\hsize]{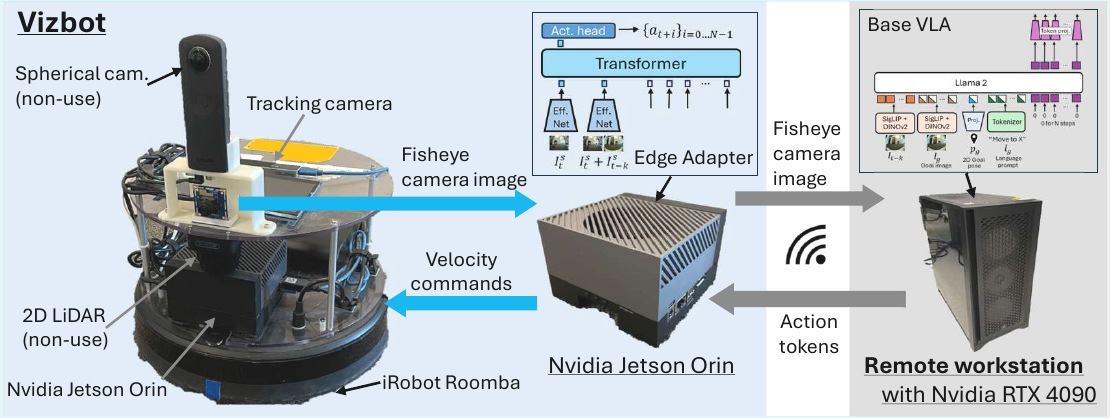}
  \end{center}
  \vspace{-2mm}
  \caption{\small \textbf{Hardware setup for our proposed system.} The \SmallModel{} is deployed on an NVIDIA Jetson Orin mounted on a mobile robot (Vizbot), while the base VLA runs on a remote workstation equipped with an NVIDIA RTX~4090.}

  \label{f:vizbot_system}
  \vspace*{-4mm}
\end{figure}
\subsection{Mobile Robot Platforms}
We evaluate \MethodName{} on Vizbot~\cite{niwa2022spatio} in Fig.~\ref{f:vizbot_system}, a low-cost mobile robot platform. The robot is equipped with multiple sensors, including a fisheye camera, a tracking camera, and wheel velocity sensors on both the left and right wheels. The external tracking camera is used to estimate the robot's pose as well as the goal location for the 2D pose-conditioned navigation tasks. At each timestep, the robot pose is expressed relative to the target pose $p_g$. 

The \SmallModel{} runs on an NVIDIA Jetson Orin configured in 30 W mode, which serves as the onboard robot controller and outputs velocity commands for navigation. The large base VLA is deployed on a workstation equipped with an NVIDIA RTX 4090 GPU. During the experiments, the action token embeddings produced by the base VLA are transmitted to the onboard robot controller via WiFi.

\begin{figure}[t]
  \begin{center}
      \includegraphics[width=0.99\hsize]{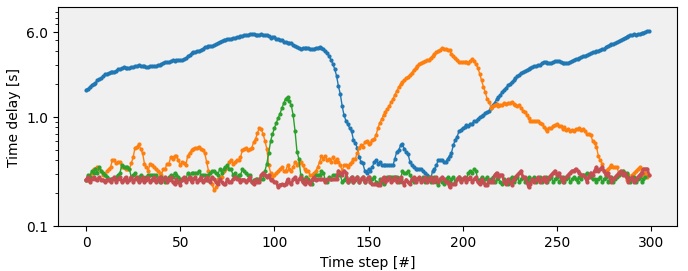}
  \end{center}
  \vspace{-2mm}
  \caption{\small \textbf{Time delay between the workstation and the robot onboard controller in our experiments.} We measure time delay across four different environments and visualize the results using distinct colors for each environment.}
  \label{f:time_delay}
  \vspace*{-4mm}
\end{figure}

We measure the interval between sending an image from the robot onboard controller and receiving the corresponding action token embedding from the workstation, corresponding to the delay step $k$ in our system. As shown in Fig.~\ref{f:time_delay}, due to WiFi conditions, the system’s time delay varies widely, ranging from 0.28 to 6.0 seconds. We evaluate the robustness of our \MethodName{} in the same environments, which include highly variable time delays.
\subsection{Baselines}
In our evaluation, we compare \MethodName{} against three state-of-the-art baselines, including lightweight policies as well as large robot foundation models. 

{{\bf OmniVLA-edge~\cite{hirose2025omnivla}:}} Since OmniVLA-edge is designed as a lightweight policy with 108M parameters, it can be executed entirely on the robot’s onboard controller at 6 Hz. Similar to OmniVLA, OmniVLA-edge supports multiple goal modalities, including 2D goal poses and language prompts.

{{\bf OmniVLA~\cite{hirose2025omnivla}:}} We evaluate the original OmniVLA, which we use as the pretrained VLA in \MethodName{}. Since OmniVLA is a large-scale policy with 8.26B parameters, we deploy it on a workstation, where it achieves a maximum inference rate of 5 Hz. To ensure smooth control, velocity commands are issued at 10 Hz by following each generated action pose chunk over 8 steps.

{{\bf Ours without E2E~\cite{sendai2025leave}:}} Since our task setting differs substantially from that of \cite{sendai2025leave}, we adopt only the core idea of their approach. We train the \SmallModel{} as a lightweight model while keeping the robotic foundation model frozen. 
During inference, similar to our method, the \SmallModel{} is deployed on the robot's onboard controller, while the base VLA runs on a remote workstation.

{{\bf Ours (workstation):}} We keep the entirety of \MethodName{} in the workstation and compare the performance with our proposed implementation using both the workstation and the robot onboard controller. This baseline demonstrates the difference in performance purely from network latency.
\section{Evaluation}
To evaluate the \MethodName{} policy, we design experiments to address the following questions:
\begin{itemize}
    \item[{\bf Q1}] Does \MethodName{} outperform VLA baselines that do not account for latencies in inference?
    \item[{\bf Q2}] Is end-to-end training important for the performance of \MethodName{}?
    \item[{\bf Q3}] Can \MethodName{} quantitatively demonstrate robustness to large delays in the navigation system?
\end{itemize}
%
%
\subsection{\MethodName{} vs. baselines}
For {\bf Q1} and {\bf Q2}, we conduct a comparative analysis with several baselines, and summarize the results in Table~\ref{tab:individual}. 

\textbf{Pose-conditioned performance.} The experiments are conducted mainly in a challenging, cluttered indoor environment, where pedestrians walk around the robot and actively interfere with its motion. Despite these conditions, \MethodName{} successfully reaches the specified goals without colliding with obstacles in 85\% of the trials, demonstrating an advantage of 40\% over the state-of-the-art baselines.

The proposed method combines high-level visual and semantic understanding from the base VLA with the \SmallModel{}, which runs at high speed on the robot’s onboard controller, enabling dynamic adaptation to disturbances. As a result, it excels at avoiding both static and dynamic obstacles. 
In contrast, ``OmniVLA'' and ``Ours (workstation)'' suffer from longer inference latencies, and the resulting delays lead to frequent collisions when running in real time. We provide the inference frequency in Table~\ref{tab:individual}, but the system also suffers from network latencies (since the large models must run off-board), which often significantly delay the actions. We characterize network latency in Fig.~\ref{f:time_delay}. The baseline methods also exhibit significant overshooting relative to the shortest path. On average, they require 15$\%$ more time to reach the goal. The OmniVLA-edge baseline can run relatively quickly on board, but because it uses a much smaller model, it has more limited visual and semantic capabilities.

Towards answering {\bf Q2}, we observe that ``Ours without E2E'', which is not trained end-to-end, fails to learn a good mapping between the base VLA and the \SmallModel{}, resulting in an overall drop in performance compared to \MethodName{}. 

\begin{figure}[t]
  \vspace{1mm}
  \begin{center}
      \includegraphics[width=0.99\hsize]{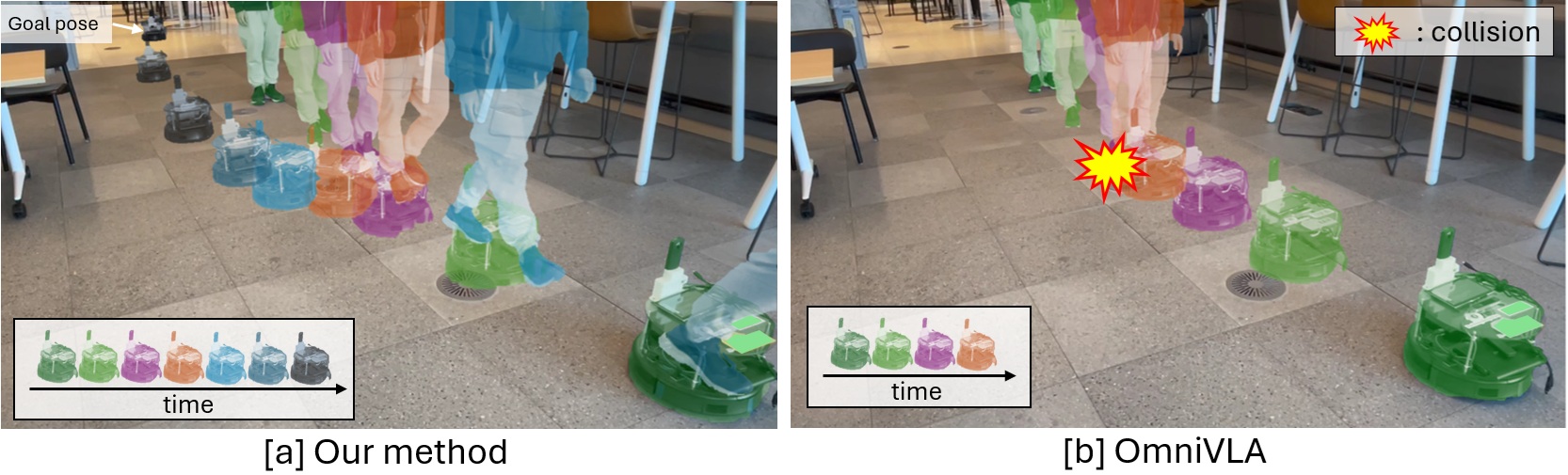}
  \end{center}
  \vspace{-2mm}
  \caption{\small {\bf Visualization of 2D pose-conditioned navigation in the presence of a pedestrian under a workstation latency of 0.2 s and fluctuating network latency.} Our \MethodName{} yields to the pedestrian and then continues along the required trajectory to reach the goal.}    
  \label{f:vis_ped}
  \vspace*{-3mm}
\end{figure}

Figure~\ref{f:vis_ped} shows a snapshot of a scenario in which a pedestrian walks toward the robot from the front. Although the robot’s maximum speed is 0.3~m/s, significantly slower than the pedestrian, Figure~\ref{f:vis_ped}[a] demonstrates that \MethodName{} allows the robot to yield to the pedestrian, pass safely without collision, and then proceed toward the goal. In contrast, ``OmniVLA'', which runs inference on a workstation, suffers from delayed action updates and consequently collides with the pedestrian, failing to reach the goal.

\begin{figure}[t]
  \vspace{1mm}
  \begin{center}
      \includegraphics[width=0.99\hsize]{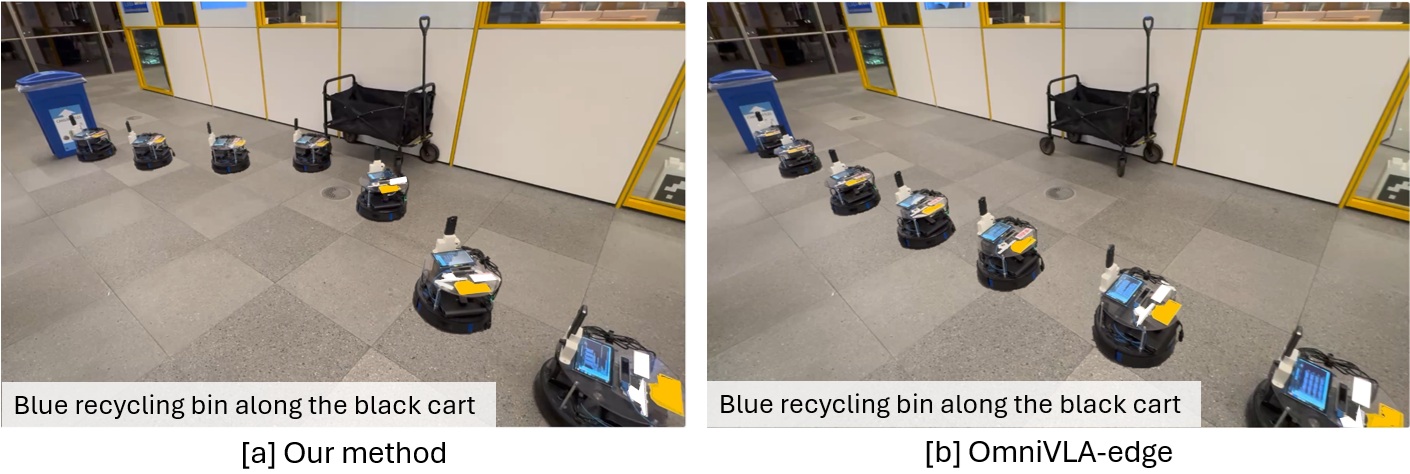}
  \end{center}
  \vspace{-2mm}
  \caption{\small {\bf Visualization of language-conditioned navigation under a workstation latency of 0.2 s.} Our \MethodName{} can achieve strong language following performance by leveraging the large base VLA.}    
  \label{f:vis_lan}
  \vspace*{-3mm}
\end{figure}

\textbf{Language-conditioned performance.} Since \MethodName{} uses OmniVLA as its pretrained model, it can leverage its ability to map high-level language and visual understanding into actions, enabling behavior that follows language instructions effectively. 
In our evaluations, \MethodName{} shows robustness to out-of-distribution language instructions, resulting in performance comparable to OmniVLA. Its performance is a substantial advantage over OmniVLA-edge, which relies on a smaller language model, as demonstrated in Fig.~\ref{f:vis_lan}. These results demonstrate that \MethodName{} effectively preserves the capabilties of OmniVLA while being reactive to its surroundings. 

\vspace{2mm}
While our evaluation system experiences delays of up to 6 seconds due to WiFi conditions, \MethodName{} consistently demonstrates robust navigation performance across all modalities. Because the \SmallModel{} is designed to leverage the most recent observation while retaining the high-level guidance of the base VLA, \MethodName{} effectively mitigates performance degradation caused by delays, guiding the robot reliably to the goal position without collision. 
%
%
\begin{table*}[t]
  \vspace{2mm}
  \begin{center}
  \resizebox{2.0\columnwidth}{!}{
  \begin{tabular}{lcccccccc} \toprule
    \textbf{Method} & & \multicolumn{2}{c}{\textbf{Implementation}} & \multicolumn{4}{c}{\textbf{2D Pose}} & \multicolumn{1}{c}{\textbf{Language}} \\
    \cmidrule(lr){3-4} \cmidrule(lr){5-8} \cmidrule(lr){9-9}
    & Model size & WS & Edge & SR$\uparrow$ & Time [s]$\downarrow$ & Static Col. [$\#$]$\downarrow$ & Dynamic Col. [$\#$]$\downarrow$ & Lan. follow$\uparrow$\\ \midrule
    OmniVLA-edge~\cite{hirose2025omnivla} & 108M & -- & \checkmark (6.0 Hz) & 0.25 & 80.07 & 0.60 & 1.00 & 0.50 \\  
    OmniVLA~\cite{hirose2025omnivla} & 8.26B & \checkmark (5.0 Hz) & --& 0.45 & 70.73 & 0.30 & 1.05 & {\bf 0.83} \\
    Ours without E2E~\cite{sendai2025leave} & 8.27B + 76M & \checkmark (5.0 Hz) & \checkmark (8.0 Hz) & 0.25 & 82.78 & 0.60 & 1.05 & 0.75 \\         
    Ours (workstation) & 8.27B + 76M & \checkmark (5.0 Hz) & --  & 0.30 & 89.79 & 0.70 & 0.50 & 0.67 \\            
    \ours Ours & 8.27B + 76M & \checkmark (5.0 Hz) & \checkmark (8.0 Hz)  & {\bf 0.85} & {\bf 59.18} & {\bf 0.10} & {\bf 0.10} & 0.75\\    
    \bottomrule
  \end{tabular}%
  }
  \end{center}
  \vspace{-2mm}
  \caption{{\bf Quantitative analysis.} ``WS'' and ``Edge'' indicates whether the implementation in running in the workstation with the Nvidia RTX 4090 or the robot edge controller with the Nvidia Jetson Orin. ``SR'' indicates the success rate, reaching the goal position without any collisions. ``Time'' indicates the time at which the robot arrives at the goal. ``Static Col.'' and ``Dynamic Col.'' indicate the average collision number with static obstacles and pedestrians, respectively. ``Behavior'' indicates the success rate in following OOD language prompts.}
  \vspace{-4mm}
  \label{tab:individual}
\end{table*}
\subsection{Quantitative analysis for time delay}
%
%
It is challenging to have reliable quantitative evaluations for the network latency, as it will naturally vary over a WiFi connection. 
To answer {\bf Q2}, we evaluate the robustness of policies inferenced on the workstation under artificially introduced time delays, simulating increased latency. Specifically, we test workstation latency of 0.2~seconds, 2.0~seconds, and 5.0~seconds, while the robot’s onboard controller computes actions at its best rate, which is independent of the environment. In this quantitative analysis, OmniVLA is used as the strongest baseline for comparison.

\begin{figure}[t]
  \vspace{-1mm}
  \begin{center}
      \includegraphics[width=0.99\hsize]{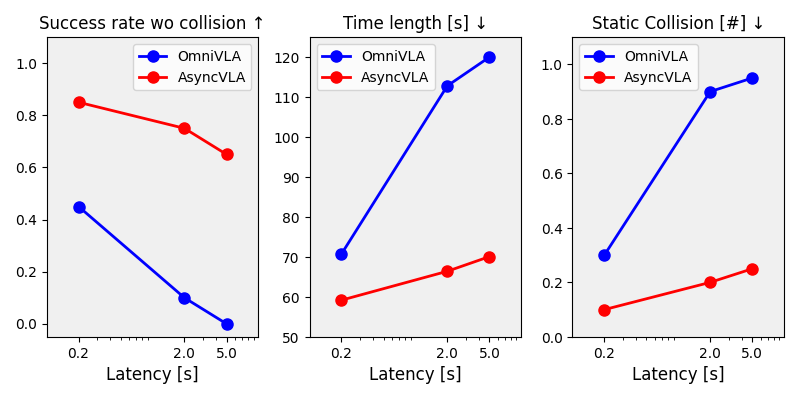}
  \end{center}
  \vspace{-4mm}
  \caption{\small {\bf Performance analysis with varying policy computation frequency in the workstation.} To evaluate robustness to time delays, we enforce artificial delays in the workstation and measure the goal arrival rate without collisions, the time taken to reach the goal pose, and the average number of collisions in 2D pose-conditioned navigation. Note that we only consider static collisions, as some methods collide immediately and fail, resulting in skewed collision numbers despite poor performance.}
  \label{f:freq_result}
  \vspace*{-5mm}
\end{figure}

Figure~\ref{f:freq_result} shows the results of 2D pose-conditioned navigation experiments under varying workstation latency, using the same metrics as in Table~\ref{tab:individual}. 
As shown in Figure~\ref{f:freq_result}, our method outperforms OmniVLA at the lowest latency of 0.2~seconds, and this performance gap increases as the artificial latency increases. As the \SmallModel{} is running on the onboard controller, it can adapt the guidance provided by the base VLA as the action token embeddings based on information provided in the current observation. 
In contrast, OmniVLA produces actions conditioned on stale observations, leading to degraded navigation performance as latencies increase.

\begin{figure}[t]
  \vspace{0mm}
  \begin{center}
      \includegraphics[width=0.99\hsize]{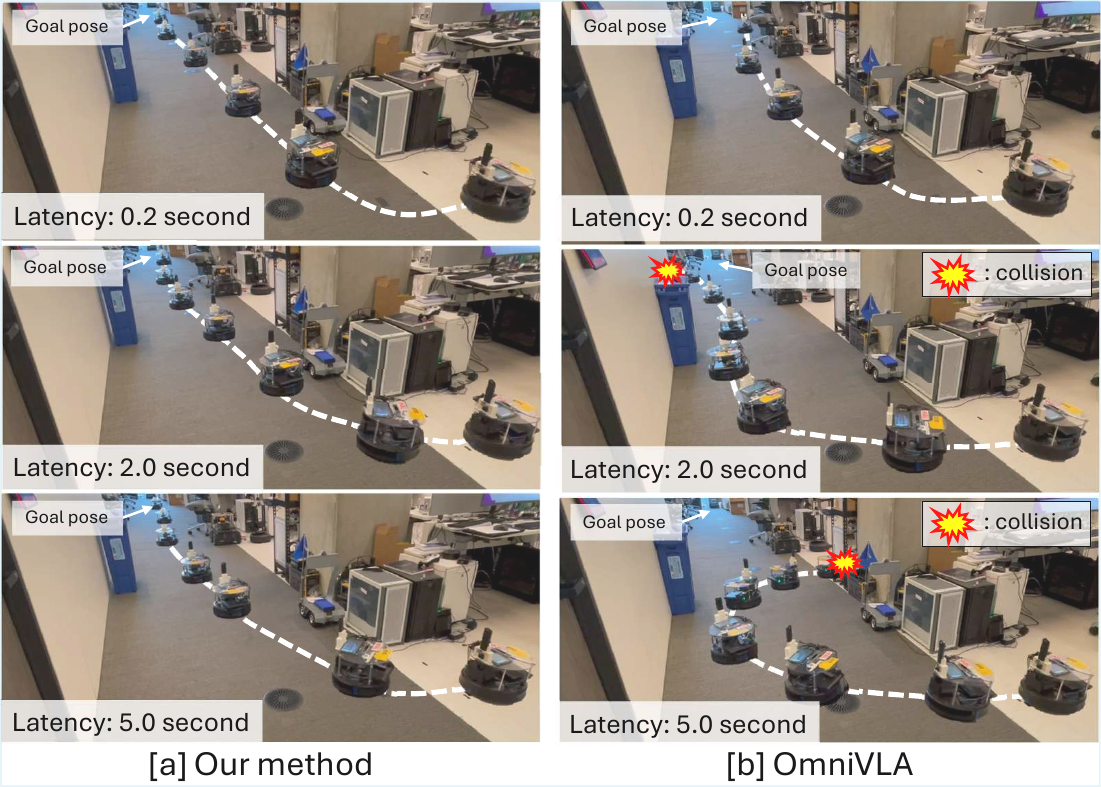}
  \end{center}
  \vspace{-2mm}
  \caption{\small {\bf Visualization of 2D pose-conditioned navigation with varying frequencies of the large model computation.} Unlike OmniVLA, our \MethodName{} consistently reaches the goal pose across different frequencies, even in cluttered environments.}
  \label{f:vis_static}
  \vspace*{-5mm}
\end{figure}

Figure~\ref{f:vis_static} shows examples of the robot's behavior at different workstation latencies. \MethodName{} demonstrates robustness to latency, maintaining similar trajectories toward the goal. In contrast, OmniVLA deviates significantly from the shortest path toward the goal pose and, in this case, collides with static obstacles, failing to reach the goal.


\section{Discussion and Future Work} 
\label{sec:conclusion}

We presented AsyncVLA, a vision- and language-conditioned navigation framework that leverages a pretrained VLA for high-level visual and semantic understanding with a lightweight \SmallModel{} running on board the robot for fast, adaptive action inference. This design enables robust navigation in dynamic and partially observable indoor environments and can effectively handle both static and dynamic obstacles while compensating for both network and inference latencies. 

We demonstrate that AsyncVLA consistently outperforms state-of-the-art baselines, including OmniVLA and OmniVLA-edge, in terms of collision avoidance, adherence to language instructions, and efficiency in goal-reaching. Notably, AsyncVLA maintains high performance even under latency up to 5 seconds, highlighting its robustness to communication latency between the workstation and robot, which are often unavoidable in real-world robot deployments. Furthermore, AsyncVLA shows strong generalization to out-of-distribution language instructions, demonstrating that \MethodName{} can still leverage the rich semantic understanding of OmniVLA while performing rapid inference with its \SmallModel{}.

Although \MethodName{} achieves strong performance in diverse navigation tasks, we are still limited to using base VLA that have open-source weights and, therefore, can be finetuned end-to-end, a step we found crucial to better align the VLA and \SmallModel{}. Additionally, \MethodName{} is limited by the small amount of dynamic interactions in available navigation datasets. As future work, we plan to explore how we can further decouple the base VLA and \SmallModel{} policies in \MethodName{} and only require training of the \SmallModel{}, which would also reduce the compute required to train \MethodName{}. Additionally, we will explore other sources of dynamic interactions that can be leveraged as training data for \MethodName{}, such as human video.

\section*{Acknowledgments}
This research was supported by Berkeley AI Research at the University of California, Berkeley and Toyota Motor North America. 
And, this work was partially support by ARL DCIST CRA W911NF-17-2-0181 and the NSF under IIS-2150826.
We thank Tatsuya Matsushima and Kyle Stachowicz for discussing the initial idea and the system architecture for \MethodName{}.
\bibliographystyle{plainnat}
\bibliography{references}
%
%
%
%
\input{appendix.tex}
\end{document}

%% file: includes.tex
\newcommand{\MethodName}{AsyncVLA}
\newcommand{\SmallModel}{Edge Adapter}

%% file: appendix.tex
\appendix
%
%
\subsection{Training Objective}
\label{app:obj}
To train our \MethodName, we calculate following $J_{im}$ and $J_{sm}$ in Eq.~\ref{eq:objective} and train our policy to minimize sum of them.
\begin{align}
J_{im} 
&= \frac{1}{N} \sum_{i=0}^{N-1} \Bigl[
(a^{\text{ref}}_{t+i} - a_{t+i})^2
+ (\Delta a^{\text{ref}}_{t+i} - \Delta a_{t+i})^2
\Bigr], \\
J_{sm} 
&= \frac{1}{N} \sum_{i=0}^{N-1} (a_{t+i+1} - a_{t+i})^2 .
\label{eq:objective_im_sm}
\end{align}
$J_{\mathrm{im}}$ denotes the imitation loss, which encourages the policy to imitate the expert action trajectories. In our design, we impose two forms of action supervision defined in different coordinate frames. Specifically, $a^{\text{ref}}_{t+i}$ represents the robot pose expressed in the robot-local coordinate frame at time $t$, while $\Delta a^{\text{ref}}_{t+i}$ represents the relative robot pose expressed with respect to the coordinate frame at the previous time step $t+i-1$. Supervision with $a^{\mathrm{ref}}_{t+i}$ is useful for learning the coarse direction in which the robot should move, while supervision with $\Delta a^{\mathrm{ref}}_{t+i}$ encourages rapid action adjustments over a few steps in response to dynamic environmental changes. Since our action head predicts action chunk $\{\Delta a_{t+i}\}_{i=0}^{N-1}$ and internally integrates them to obtain $\{a_{t+i}\}_{i=0}^{N-1}$, we can compute $J_{im}$ using these actions. $J_{sm}$ is the objective to minimize the action deltas for regularization. In $J_{sm}$, we set $a_{t-1}$ as the origin. For the LeLaN dataset, since no raw action references are available, we penalize only the final generated action $a_{t+N-1}$ to encourage proximity to the target object pose specified by the language prompt, following the LeLaN policy training procedure.
\subsection{Inference details}
\label{app:inf}
\MethodName{} is implemented using Robot Operating System (ROS). The communication pipeline begins by transmitting the observed image $I_t$ from the robot's onboard controller to a workstation as a compressed image topic. Each image is appended with a timestamp at the moment of capture to ensure temporal synchronization. On the workstation, the base VLA processes the observation and transmits the resulting action token embeddings $\{a^{\text{emb}}_{t+i}\}_{i=0 \ldots N-1}$ back to the robot via a ``VectorStamped'' topic. This return message retains the original timestamp of the image $I_{t-k}$ to allow the robot edge controller to identify the specific observation used for inference. To facilitate this, we maintain a buffer of observed images and their corresponding timestamps, enabling the system to retrieve the matching $I_{t-k}^s$ upon receiving the embeddings.

The edge controller processes the retrieved $I_{t-k}^s$ alongside the most recent observation $I_{t}^s$ to generate an action pose chunk $\{a_{t+i}\}_{i=0}^{N-1}$. Following established protocols in the literature~\citep{shahvint, sridhar2024nomad, shah2023gnm}, we employ a PD controller to reach the future waypoint $a_{t+2}$, from which we derive the linear and angular velocity commands. To ensure a fair comparison with baseline methods that exhibit lower inference frame rates, we utilize a PD controller to follow the waypoint trajectory $\{a_{t+i}\}_{i=0}^{N-1}$ and publish velocity commands at a consistent frequency of 10.0 Hz. For these baselines, the PD controller updates the active waypoint trajectory whenever a new output is received from the baseline policy. Across all evaluated methods, velocity commands are issued as "Twist" topics to drive the mobile robot.
\subsection{Evaluation details}
\label{app:eval}
To evaluate goal pose-conditioned navigation, we selected ten challenging environments, including cluttered indoor spaces and outdoor areas with weak Wi-Fi connectivity. We conducted two trials at different times in each environment to ensure variety. In one of these trials, pedestrians were present and engaged in three specific types of interactions with the robot. These interactions, illustrated in the supplemental videos, were kept consistent across all evaluated methods to ensure a fair and reproducible comparison.

We define a successful trial as one where the robot reaches a position within 1 meter of the goal without any collisions and within a 120-second time limit. In instances where the robot is able to continue moving after a collision with static obstacles or dynamic pedestrians, navigation is maintained, and the total time taken to reach the goal is recorded. Conversely, if the robot fails to reach the goal, a constant value of 120 seconds is assigned to the trial for the purpose of calculating average performance. Navigation is only manually terminated if the time exceeds 120 seconds or if the robot enters a potentially unsafe situation, such as approaching a descending staircase.
\subsection{Breakdown of quantitative analysis}
\label{app:break}
Table~\ref{tab:individual_apx} provides a detailed breakdown of our quantitative experiments for goal pose-conditioned navigation, with metrics categorized by trials conducted with and without pedestrians. The strongest baseline, OmniVLA operating at a 5~Hz inference rate, performs effectively in static scenes, reaching the goal without collision in 90$\%$ of trials. However, its performance significantly degrades in dynamic environments, where the latency between the edge controller and the workstation becomes critical. Furthermore, as discussed in the main paper, OmniVLA exhibits a drastic drop in performance when the frame rate is reduced. In contrast, our \MethodName{} maintains high performance in static scenes even at low frame rates. Because our \SmallModel{} reacts rapidly to current observations, \MethodName{} successfully avoids collisions and reaches the goal position more reliably. While performance decreases slightly as the workstation's frame rate is reduced, our \MethodName{} with 0.2~Hz inference still successfully guides the robot to the goal in 50$\%$ of the cases.
%
%
\begin{table*}[t]
  \vspace{2mm}
  \begin{center}
  \resizebox{2.0\columnwidth}{!}{
  \begin{tabular}{lcccccccc} \toprule
    \textbf{Method} & & & \multicolumn{2}{c}{\textbf{Implementation}} & \multicolumn{4}{c}{\textbf{2D Pose}} \\
    \cmidrule(lr){4-5} \cmidrule(lr){6-9}
    & Model size & Pedestrian & WS & Edge & SR$\uparrow$ & Time [s]$\downarrow$ & Static Col. [$\#$]$\downarrow$ & Dynamic Col. [$\#$]$\downarrow$ \\ \midrule
    OmniVLA-edge~\cite{hirose2025omnivla} & 108M & & -- & \checkmark (6.0 Hz) & 0.50 & 76.36 & 0.50 & -- \\ 
    & 108M & \checkmark & -- & \checkmark (6.0 Hz) & 0.00 & 83.77 & 0.70 & 2.00 \\      
    OmniVLA~\cite{hirose2025omnivla} & 8.26B & & \checkmark (5.0 Hz) & --& 0.90 & 60.36 & 0.10 & -- \\
    & 8.26B & \checkmark & \checkmark (5.0 Hz) & --& 0.00 & 81.09 & 0.50 & 2.10 \\   
    OmniVLA~\cite{hirose2025omnivla} & 8.26B & & \checkmark (0.5 Hz) & --& 0.10 & 112.81 & 0.90 & -- \\
    & 8.26B & \checkmark & \checkmark (0.5 Hz) & --& 0.00 & 112.79 & 0.90 & 1.60 \\  
    OmniVLA~\cite{hirose2025omnivla} & 8.26B & & \checkmark (0.2 Hz) & --& 0.00 & 120.00 & 1.00 & -- \\
    & 8.26B & \checkmark & \checkmark (0.2 Hz) & --& 0.00 & 120.00 & 0.90 & 1.70 \\      
    Ours without E2E~\cite{sendai2025leave} & 8.27B + 76M & & \checkmark (5.0 Hz) & \checkmark (8.0 Hz) & 0.50 & 74.98 & 0.50 & -- \\ 
    & 8.27B + 76M & \checkmark & \checkmark (5.0 Hz) & \checkmark (8.0 Hz) & 0.00 & 90.57 & 0.70 & 2.10 \\        
    Ours (workstation) & 8.27B + 76M & & \checkmark (5.0 Hz) & --  & 0.40 & 81.83 & 0.60 & -- \\   
    & 8.27B + 76M & \checkmark & \checkmark (5.0 Hz) & --  & 0.20 & 97.75 & 0.80 & 1.00 \\       
    \ours Ours & 8.27B + 76M & & \checkmark (5.0 Hz) & \checkmark (8.0 Hz)  & 0.90 & 52.95 & 0.10 & -- \\    
    \ours & 8.27B + 76M & \checkmark & \checkmark (5.0 Hz) & \checkmark (8.0 Hz)  & 0.80 & 65.41 & 0.10 & 0.20 \\   
    \ours Ours & 8.27B + 76M & & \checkmark (0.5 Hz) & \checkmark (8.0 Hz)  & 0.90 & 61.11 & 0.10 & -- \\    
    \ours & 8.27B + 76M & \checkmark & \checkmark (0.5 Hz) & \checkmark (8.0 Hz)  & 0.60 & 71.71 & 0.30 & 0.30 \\  
    \ours Ours & 8.27B + 76M & & \checkmark (0.2 Hz) & \checkmark (8.0 Hz)  & 0.80 & 67.83 & 0.20 & -- \\    
    \ours & 8.27B + 76M & \checkmark & \checkmark (0.2 Hz) & \checkmark (8.0 Hz)  & 0.50 & 72.33 & 0.30 & 0.70 \\      
    \bottomrule
  \end{tabular}%
  }
  \end{center}
  \vspace{0mm}
  \caption{{\bf Breakdown of quantitative analysis.} ``WS'' and ``Edge'' indicate whether the implementation is running in the workstation with the Nvidia RTX 4090 or the robot edge controller with the Nvidia Jetson Orin. ``SR'' indicates the success rate, reaching the goal position without any collisions. ``Time'' indicates the time at which the robot arrives at the goal. ``Static Col.'' and ``Dynamic Col.'' indicate the average collision number with static obstacles and pedestrians, respectively. ``Behavior'' indicates the success rate in following OOD language prompts.}
  \vspace{0mm}
  \label{tab:individual_apx}
\end{table*}
\begin{figure*}[t]
    \centering
    \begin{subfigure}{0.32\linewidth}
        \centering
        \includegraphics[width=\linewidth]{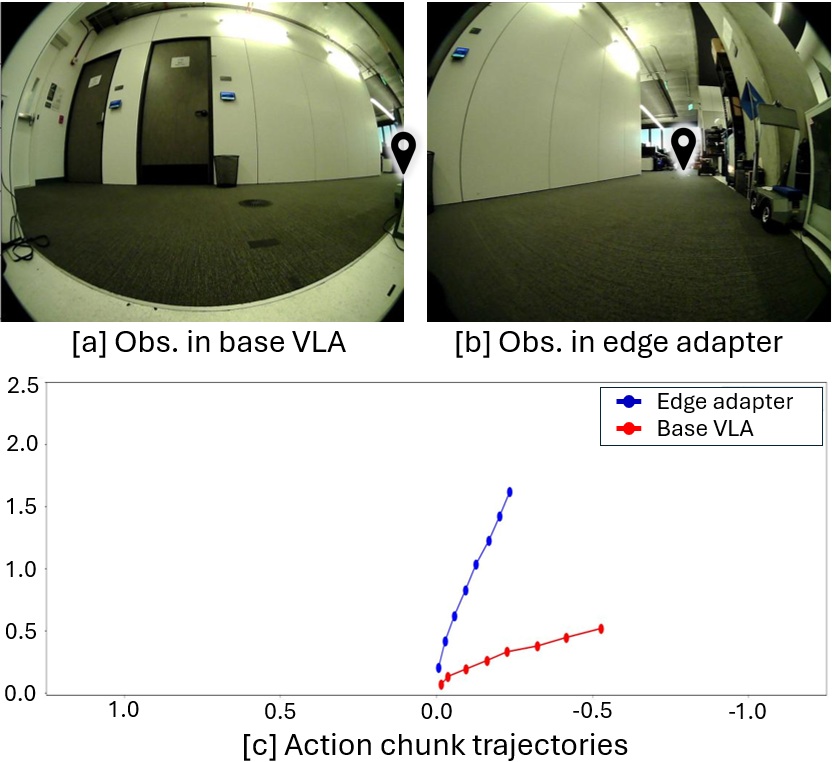}
        \caption{Case A}
    \end{subfigure}
    \hfill
    \begin{subfigure}{0.32\linewidth}
        \centering
        \includegraphics[width=\linewidth]{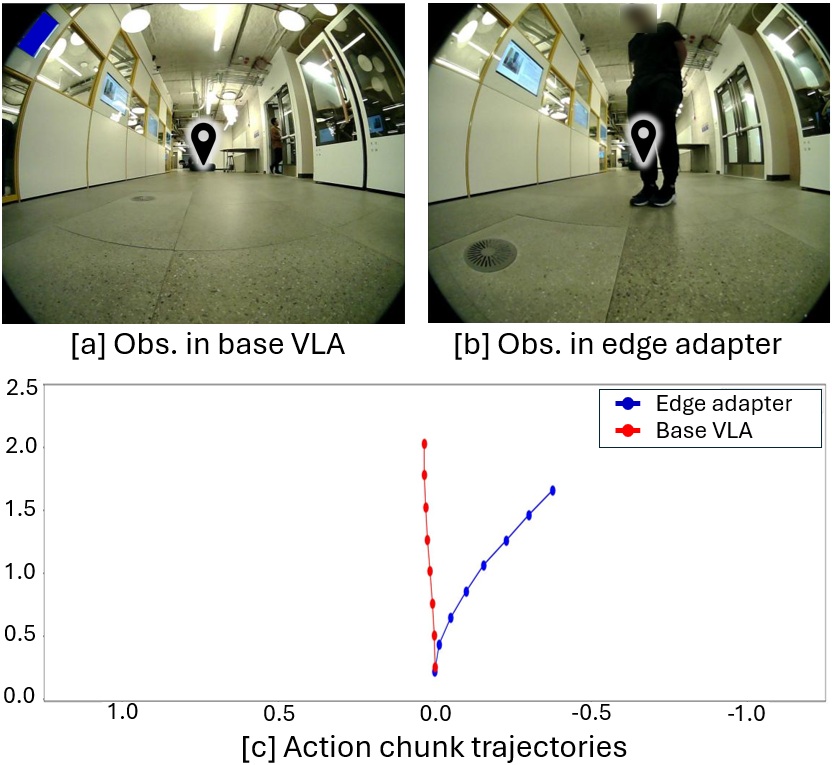}
        \caption{Case B}
    \end{subfigure}
    \hfill
    \begin{subfigure}{0.32\linewidth}
        \centering
        \includegraphics[width=\linewidth]{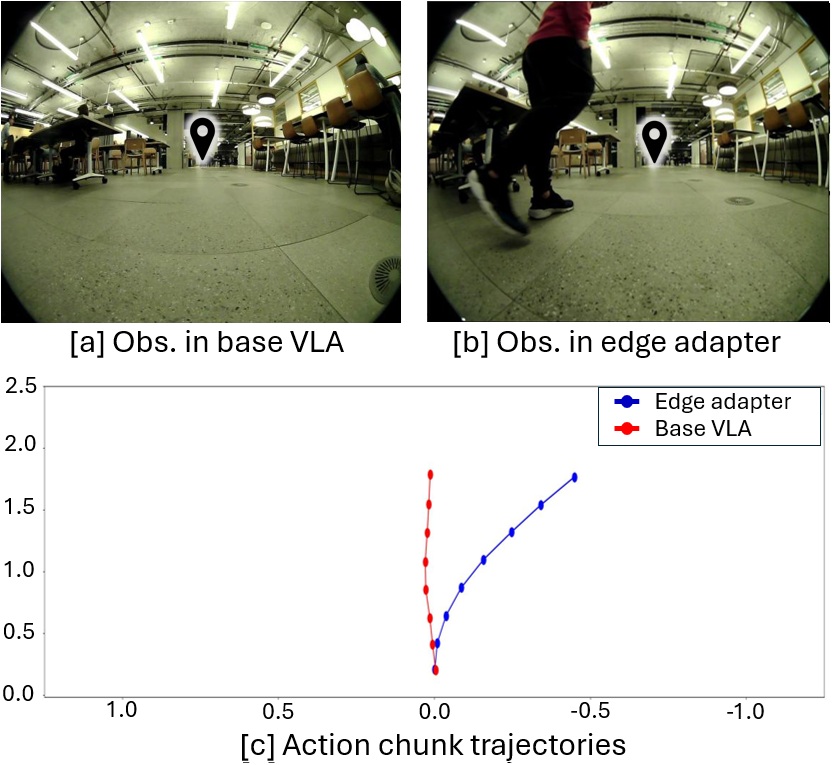}
        \caption{Case C}
    \end{subfigure}
    \caption{{\bf Visualization of generated action chunks from \MethodName{}.} We present three examples of the generated action chunks during navigation. Panel (a) shows the observation provided to the base VLA, denoted as $I_{t-k}$, while panel (b) shows the observation provided to the edge adapter, denoted as $I_t$. Panel (c) visualizes the corresponding generated action chunks. In panel (c), the blue curve represents the action chunk generated by the edge adapter using the current observation $I_t$, whereas the red curve represents the action chunk obtained by feeding the delayed observation $I_{t-k}$ into both the base VLA and the edge adapter. We refer to this red curve as the action chunk of the base VLA, as it is conditioned on the observation $I_{t-k}$.}
    \label{f:vis_act}
    \vspace{-4mm}
\end{figure*}
\subsection{Visualization of action chunk from \MethodName{}}
\label{app:vis_act}
We visualize the generated action chunks from our \MethodName{} in Fig.~\ref{f:vis_act}. We sample three scenes from our experiments and show the observation provided to the base VLA, $I_{t-k}$, in [a], the observation provided to the edge adapter, $I_t$, in [b], and the corresponding action chunk trajectories in [c]. In [c], we plot two trajectories: the blue curves represent the action chunks generated by the edge adapter (our method), while the red curves represent the action chunks obtained by feeding the delayed observation $I_{t-k}$ into both the base VLA and the edge adapter, which we treat as the action chunks of the base VLA. Intuitively, the base VLA provides features to the edge adapter to generate the red trajectories. Across all scenes, the edge adapter appropriately deviates from the red trajectories and refines them to move toward the goal pose in [i] and to avoid collisions with pedestrians in [ii] and [iii]. In [i], the edge adapter correctly infers the relative pose between $I_t$ and $I_{t-k}$ and generates actions toward the target goal. In [ii] and [iii], although pedestrians do not appear in the observations provided to the base VLA, the edge adapter reacts to the current observations, which do capture pedestrians, and generates collision-avoidance behaviors.